\documentclass{article} 
\usepackage{nips15submit_e}
\usepackage{hyperref}
\usepackage{url}
\usepackage{graphicx}
\usepackage{amsmath, dsfont, amsfonts}
\usepackage[numbers]{natbib}
\usepackage{color}

\title{Model Selection in Bayesian Neural Networks via Horseshoe Priors}
\author{
Soumya Ghosh \\
{IBM T.J. Watson Research Center}\\
\texttt{ghoshso@us.ibm.com} \\
\and
\textbf{Finale Doshi-Velez} \\
{Harvard University} \\
\texttt{finale@seas.harvard.edu} \\
}

\nipsfinalcopy

\begin{document}

\maketitle

\newcommand{\Xmat}{\text{X}}
\newcommand{\Ymat}{\text{Y}}
\newcommand{\W}[1]{\mathcal{W}_{#1}}
\newcommand{\s}{\mathcal{S}}
\newcommand{\ind}[1]{\mathbf{1}[#1]}
\newcommand{\wvar}[1]{\tau_#1^{-1}}
\newcommand{\data}{\mathcal{D}}
\newcommand{\eye}{\mathbb{I}}
\newcommand{\T}{\mathcal{\theta}}
\newcommand{\Tau}{\mathcal{T}}
\newcommand{\blayer}{b_g}
\newcommand{\bnode}{b_0}
\newcommand{\invgamma}{\text{Inv-Gamma}}
\newcommand{\normal}{\mathcal{N}}
\newcommand{\halfcauchy}{C^{+}}
\newcommand{\taunode}{\tau_{kl}}
\newcommand{\taulayer}{\upsilon_{l}}
\newcommand{\lambdanode}{\lambda_{kl}}
\newcommand{\lambdalayer}{\vartheta_{l}}
\newcommand{\lambdakappa}{\rho_{\kappa}}
\newcommand{\unode}{u_{kl}}
\newcommand{\wnode}{w_{kl}}
\newcommand{\betanode}{\beta_{kl}}
\newcommand{\half}{\frac{1}{2}}
\newcommand{\elbo}{\mathcal{L}}
\newcommand{\gradelbo}{\nabla_\phi\hat{\mathcal{L}}(\phi)}
\newcommand{\E}[1]{\mathbb{E}[#1]}
\newcommand{\Ewrt}[2]{\mathbb{E}_{#1}[#2]}
\newcommand{\ent}[1]{\mathbb{H}[#1]}
\newcommand{\real}[1]{\mathbb{R}^{#1}}
\newcommand{\bkappa}{b_\kappa}

\begin{abstract}
Bayesian Neural Networks (BNNs) have recently received increasing attention for their ability to provide well-calibrated posterior uncertainties.  However, model selection---even choosing the number of nodes---remains an open question.  In this work, we apply a horseshoe prior over node pre-activations of a Bayesian neural network, which effectively turns off nodes that do not help explain the data.  We demonstrate that our prior prevents the BNN from under-fitting even when the number of nodes required is grossly over-estimated.  Moreover, this model selection over the number of nodes doesn't come at the expense of predictive or computational performance; in fact, we learn smaller networks with comparable predictive performance to current approaches.
\end{abstract}

\section {Introduction}
Bayesian Neural Networks (BNNs) are increasingly the de-facto approach
for modeling stochastic functions.  By treating the weights in a
neural network as random variables, and performing posterior inference
on these weights, BNNs can avoid overfitting in the regime of small
data, provide well-calibrated posterior uncertainty estimates, and
model a large class of stochastic functions with heteroskedastic and
multi-modal noise.  These properties have resulted in BNNs being
adopted in applications ranging from active learning~\cite{MHLobato15,
  YGal16Active} and reinforcement learning~\cite{CBlundell15}.

While there have been many recent advances in training
BNNs~\cite{MHLobato15, CBlundell15, DRezende14, CLouizos16,
  MHLobato16}, model-selection in BNNs has received relatively
less attention.  Unfortunately, the consequences for a poor choice of
architecture are severe: too few nodes, and the BNN will not be
flexible enough to model the function of interest; too many nodes, and
the BNN predictions will have large variance because the posterior
uncertainty in the weights will remain large.  In other approaches to
modeling stochastic functions such as Gaussian Processes (GPs), such
concerns can be addressed via optimizing continuous kernel parameters;
In BNNs, the number of nodes in a layer is a discrete
quantity. Practitioners typically perform model-selection via onerous
searches over different layer sizes.

In this work, we demonstrate that we can perform
computationally-efficient and statistically-effective model selection
in Bayesian neural networks by placing Horseshoe (HS)
priors~\cite{CCarvalho09} over the variance of weights incident to each
node in the network.  The HS prior has heavy tails and supports both
zero values and large values.  Fixing the mean of the incident weights
to be zero, nodes with small variance parameters are effectively
turned off---all incident weights will be close to zero---while nodes
with large variance parameters can be interpreted as active.  In this
way, we can perform model selection over the number of nodes required
in a Bayesian neural network.

While they mimic a spike-and-slab approach that would assign a
discrete on-off variable to each node, the continuous relaxation
provided by the Horseshoe prior keeps the model differentiable; with
appropriate parameterization, we can take advantage of recent advances
in variational inference (e.g.~\cite{DKingma14}) for training.  We
demonstrate that our approach avoids under-fitting even with the
required number of nodes in the network is grossly over-estimated; we
learn compact network structures without sacrificing---and sometimes
improving---predictive performance.

\section {Bayesian Neural Networks}
A deep neural network with $L-1$ hidden layers is parameterized by a
set of weight matrices $\W{} = \{W_l\}_{1}^{L}$, with each weight
matrix $W_l$ being of size $\real{K_l + 1\times K_{l+1}}$ where $K_l$
is the number of units in layer $l$.  The neural network maps an input
$x \in \real{D}$ to a response $f(\W{}, x)$ by recursively applying
the transformation $h(W_l[z_l, 1]^T)$, where the vector $z_l$ is
the input into layer $l$, the initial input $z_0 = x$, and $h$ is
some point-wise non-linearity, for instance the rectified-linear
function, $h(a) = \text{max}(0, a)$.

A Bayesian neural network captures uncertainty in the weight
parameters $\W{}$ by endowing them with distributions $\W{} \sim
p(\W{})$. Given a dataset of $N$ observation-response pairs $\data =
\{x_n, y_n\}_{n=1}^{N}$, we are interested in estimating the posterior
distribution,
\begin{equation}
p(\W{}\mid \data) = \frac{\prod_{n=1}^N p(y_n\mid f(\W{}, x_n))p(\W{})}{p(\data)},
\end{equation}
and leveraging the learned posterior for predicting responses to unseen
data $x_*$,
$p(y_* \mid x_*) = \int p(y_*\mid f(\W{},x_*))p(\W{}\mid\data) d\W{}$.	
The prior $p(\W{})$ allows one to encode problem-specific beliefs as well as general properties about weights.  In the past, authors have used fully factorized Gaussians~\cite{MHLobato15} on each weight $w_{k'kl}$,  structured Gaussians~\cite{CLouizos16, YGal16} on each layer $W_l$ as well as a two component scale mixture of Gaussians~\cite{CBlundell15} on each weight $w_{k'kl}$. The scale mixture prior has been shown to encourage weight sparsity. In this paper, we show that by using carefully constructed infinite scale mixtures of Gaussians, we can induce heavy-tailed priors over network weights. Unlike previous work, we force all weights incident into a unit to share a common prior allowing us to induce sparsity at the unit level and prune away units that do not help explain the data well.   

\section{Automatic Model Selection through Horseshoe Priors}
\begin{figure}[t]
\centering 
\includegraphics[width=0.75\textwidth]{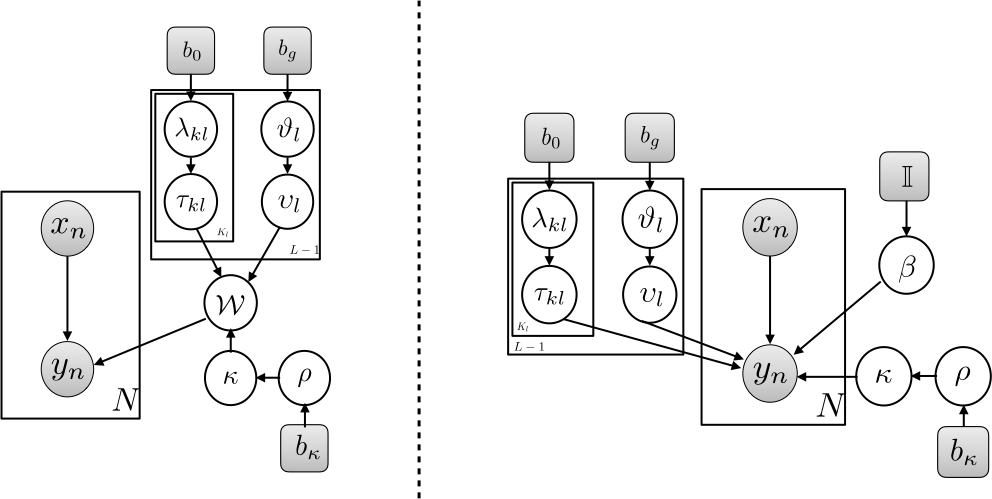}
\caption{\small{Graphical models capturing the conditional dependencies assumed by Bayesian Neural Networks with Horseshoe priors. Left: Centered parameterization, Right: Non-centered parameterization necessary for robust inference.}}	
\label{fig:gm}
\end{figure}
Let the node weight vector $\wnode \in \real{K_{l-1}+1}$ denote the set of all weights incident into unit $k$ of hidden layer $l$. We assume that each node weight vector $\wnode$ is conditionally independent and distributed according to a Gaussian scale mixture,
\begin{equation}
\wnode \mid \taunode, \taulayer \sim \mathcal{N}(0, (\taunode^2\taulayer^2)\eye),  \quad \taunode \sim \halfcauchy(0, \bnode), \quad \taulayer \sim \halfcauchy(0, \blayer).
\label{eq:HS}
\end{equation}
Here, $\eye$ is an identity matrix, $a \sim \halfcauchy(0, b)$ is the Half-Cauchy distribution with density $p(a|b) = 2/\pi b(1 + (a^2/b^2))$ for $a>0$, $\taunode$ is a unit specific scale parameter, while the scale parameter $\taulayer$ is shared by all units in the layer.

The distribution over weights in Equation~\ref{eq:HS} is called horseshoe prior~\cite{CCarvalho09}. It exhibits Cauchy-like flat, heavy tails while maintaining an infinitely tall spike at zero. Consequently, it has the desirable property of allowing sufficiently large node weight vectors $\wnode$ to escape un-shrunk---by having a large scale parameter---while providing severe shrinkage to smaller weights. This is in contrast to Lasso style regularizers and their Bayesian counterparts that provide uniform shrinkage to all weights. By forcing the weights incident on a unit to share scale parameters, the prior in Equation~\ref{eq:HS} induces sparsity at the unit level, turning off units that are unnecessary for explaining the data well. Intuitively, the shared layer wide scale $\taulayer$ pulls all units in layer $l$ to zero, while the heavy tailed unit specific $\taunode$ scales allow some of the units to escape the shrinkage. 

\paragraph{Parameterizing for More Robust Inference: Decomposing the Cauchy Distribution}
While a direct parameterization of the Half-Cauchy distribution in Equation~\ref{eq:HS} is possible, it leads to challenges during variational learning. Standard exponential family variational approximations struggle to capture the thick Cauchy tails, while a Cauchy approximating family leads to high variance gradients. Instead, we use a more convenient auxiliary variable parameterization~\cite{MWand11},
\begin{equation}
a \sim \halfcauchy(0, b)	\Longleftrightarrow a \mid \lambda \sim \invgamma(\frac{1}{2}, \frac{1}{\lambda}); \quad \lambda \sim \invgamma(\frac{1}{2}, \frac{1}{b^2}),
\end{equation}
where $v \sim \invgamma(a ,b)$ is the Inverse Gamma distribution with density $p(v) \propto v^{-a-1}\text{exp}\{-b/v\}$ for $v>0$. Since the number of output units is fixed by the problem at hand, there is no need for sparsity inducing prior for the output layer. We place independent Gaussian priors, $w_{kL} \sim \mathcal{N}(0, \kappa^2\eye)$ with vague hyper-priors $\kappa \sim \halfcauchy(0, \bkappa=5)$ on the output layer weights.  

The joint distribution of the Horseshoe Bayesian neural network is then given by,
\begin{equation}
\begin{split}
p(\data, \theta) = r(\kappa, \lambdakappa \mid \bkappa)&\prod_{k=1}^{K_L}\mathcal{N}(w_{kL} \mid 0, \kappa\eye) 
\prod_{l=1}^{L} r(\taulayer, \lambdalayer \mid \blayer) \\
&\prod_{k=1}^{K_l}r(\taunode, \lambdanode \mid \bnode)\normal(\wnode\mid 0, (\taunode^2\taulayer^2)\eye) 
\prod_{n=1}^N p(y_n\mid f(\W{}, x_n)),	
\end{split}
\end{equation}
where $ p(y_n\mid f(\W{}, x_n))$ is an appropriate likelihood function, and, $r(a, \lambda \mid b) =\invgamma(a \mid 1/2, 1/\lambda)\invgamma(\lambda \mid 1/2, 1/b^2)$,
 with $\T = \{\W{}, \mathcal{T}, \kappa, \lambdakappa\}$, $\mathcal{T} =\{ \{\taunode\}_{k=1,l=1}^{K,L}, \{\taulayer\}_{l=1}^L, \{\lambdanode\}_{k=1,l=1}^{K,L}, \{\lambdalayer\}_{l=1}^L\}$. 

\paragraph{Parameterizing for More Robust Inference: Non-Centered Parameterization}
The horseshoe prior of Equation~\ref{eq:HS} exhibits strong correlations between the  weights $\wnode$ and the scales $\taunode\taulayer$. Indeed, its favorable sparsity inducing properties stem from this coupling. However, an unfortunate consequence is a strongly coupled posterior that exhibits pathological funnel shaped geometries~\cite{MBetancourt15, JIngraham16} and is difficult to reliably sample or approximate. Fully factorized approximations are particularly problematic and can lead to non-sparse solutions erasing the benefits of using the horseshoe prior. 

Recent work~\cite{MBetancourt15, JIngraham16} suggests that the problem can be alleviated by adopting non-centered parameterizations. Consider a reformulation of Equation~\ref{eq:HS},    
\begin{equation}
\beta_{kl} \sim \normal(0, \eye)	, \quad \wnode = \taunode\taulayer\beta_{kl},
\end{equation}
where the distribution on the scales are left unchanged. Such a parameterization is referred to as non-centered since the scales and weights are sampled from independent prior distributions and are \emph{marginally} uncorrelated. The coupling between the two is now introduced by the likelihood, when conditioning on observed data. Non-centered parameterizations are known to lead to simpler posterior geometries~\cite{MBetancourt15}. Empirically, we find that adopting a non-centered parameterization significantly improves the quality of our posterior approximation and helps us better find sparse solutions. Figure~\ref{fig:gm} summarizes the conditional dependencies assumed by the centered and the non-centered Horseshoe Bayesian neural networks model.

%
%
%

\section{Learning Bayesian Neural Networks with Horseshoe priors}
We use variational inference to approximate the intractable posterior $p(\T \mid \data)$. By exploiting recently proposed stochastic extensions we are able to scale to large architectures and datasets, and deal with non-conjugacy. 
We proceed by selecting a tractable family of distributions $q(\T \mid \phi)$, with free variational parameters $\phi$. We then optimize $\phi$ such that the Kullback-Liebler divergence between the approximation and the true posterior, $\text{KL}(q(\theta\mid \phi) || p(\theta \mid \data))$ is minimized.  This is equivalent to maximizing the lower bound to the marginal likelihood (or evidence) $p(\data)$, $p(\data) \geq \elbo(\phi) = \Ewrt{q_\phi}{\text{ln } p (\data, \theta)} + \ent{q(\theta\mid\phi)}$.	
\\\\
\textbf{Approximating Family}
We use a fully factorized variational family, 
\begin{equation}
\small
\begin{split}
q(\T \mid \phi) = q(\kappa\mid\phi_\kappa)q(\lambdakappa\mid\phi_{\lambdakappa})\prod_{i,j,l}q(\beta_{ij,l}\mid \phi_{\beta_{ij,l}}) &\prod_{k,l} q(\taunode\mid\phi_{\taunode})q(\lambdanode\mid\phi_{\lambdanode})\\
&\prod_l q(\taulayer\mid\phi_{\taulayer})q(\lambdalayer\mid\phi_{\lambdalayer}).
\end{split}
\end{equation}
We restrict the variational distribution for the non-centered weight $\beta_{ij,l}$ between units $i$ in layer $l-1$ and $j$ in layer $l$, $q(\beta_{ij,l}\mid \phi_{\beta_{ijl}})$ to the Gaussian family $\normal(\beta_{ij,l}\mid \mu_{ij,l},\sigma^2_{ij,l})$. We will use $\beta$ to denote the set of all non-centered weights in the network. The non-negative scale parameters $\taunode$ and $\taulayer$ and the variance of the output layer weights are constrained to the log-Normal family, $q(\text{ln } \taunode\mid \phi_{\taunode})  = \normal(\mu_{\taunode}, \sigma^2_{\taunode})$, $q(\text{ln } \taulayer \mid \phi_{\taulayer}) = \normal(\mu_{\taulayer}, \sigma^2_{\taulayer})$, and $q(\text{ln } \kappa \mid \phi_{\kappa}) = \normal(\mu_{\kappa}, \sigma^2_{\kappa})$. We do not impose a distributional constraint on the variational approximations of the auxiliary variables $\lambdalayer$, $\lambdanode$, or $\lambdakappa$, but we will see that conditioned on the remaining variables the optimal variational family for these latent variables follow inverse Gamma distributions.
%
\\\\
\textbf{Evidence Lower Bound}
The resulting evidence lower bound (ELBO),
\begin{equation}
\small
	\elbo(\phi) = \sum_n \E{\text{ln } p(y_n \mid f(\beta, \Tau, \kappa, x_n))} + 
	\E{\text{ln } p(\Tau, \beta, \kappa, \lambdakappa \mid \bnode, \blayer, \bkappa)} +  \ent{q(\theta\mid \phi)},
\label{eq:elbo}
\end{equation}
is challenging to evaluate. The non-linearities introduced by the neural network and the potential lack of conjugacy between the neural network parameterized likelihoods and the Horseshoe priors render the first expectation in Equation~\ref{eq:elbo} intractable. Consequently, the traditional prescription of optimizing the ELBO by cycling through a series of fixed point updates is no longer available. 

\subsection{Black Box Variational Inference} 
Recent progress in black box variational inference~\cite{DKingma14, DRezende14, RRanganath14, MTitsias14} provides a recipe for subverting this difficulty. These techniques provide noisy unbiased estimates of the gradient $\gradelbo$, by approximating the offending expectations with unbiased Monte-Carlo estimates and relying on either score function estimators~\cite{RWilliams92, RRanganath14} or reparameterization gradients~\cite{DKingma14, DRezende14, MTitsias14} to differentiate through the sampling process. With the unbiased gradients in hand, stochastic gradient ascent can be used to optimize the ELBO. In practice, reparameterization gradients exhibit significantly lower variances than their score function counterparts and are typically favored for differentiable models. The reparameterization gradients rely on the existence of a parameterization that separates the source of randomness from the parameters with respect to which the gradients are sought. For our Gaussian variational approximations, the well known non-centered parameterization, $\zeta \sim \mathcal{N}(\mu, \sigma^2) \Leftrightarrow \epsilon \sim \mathcal{N}(0, 1), \zeta = \mu + \sigma \epsilon$,  allows us to compute Monte-Carlo gradients,
\begin{equation}
\small
\begin{split}
\nabla_{\mu, \sigma} \mathbb{E}_{q_w}[g(w)] \Leftrightarrow \nabla_{\mu, \sigma} \mathbb{E}_{\mathcal{N}(\epsilon\mid0,1)}[g(\mu + \sigma \epsilon)] 
 &= \mathbb{E}_{\mathcal{N}(\epsilon\mid0,1)} [\nabla_{\mu, \sigma} g(\mu + \sigma \epsilon)] \\
 &= \frac{1}{S} \sum_s \nabla_{\mu, \sigma} g(\mu + \sigma \epsilon^{(s)}),\end{split}
\label{eq:reparam}
\end{equation}
for any differentiable function $g$ and $\epsilon^{(s)} \sim \mathcal{N}(0,1)$.
Further, as shown in~\cite{DKingma15}, the variance of the gradient estimator can be provably lowered by noting that the weights in a layer only affect $\elbo(\phi)$ through the layer's pre-activations and directly sampling from the relatively lower-dimensional variational posterior over pre-activations. 

\paragraph{Variational distribution on pre-activations} Recall that the pre-activation of node $k$ in layer $l$, $\unode$ in our non-centered model is $\unode = \taunode\taulayer\betanode^T [a, 1]^T$. The variational posterior for the pre-activations is given by,
\begin{equation}
\begin{split}
q(\unode\mid \mu_{\unode}, \sigma^2_{\unode}) = \normal(\unode\mid \mu_{\unode}, \sigma^2_{\unode}), \\
\mu_{\unode} = \mu_{\betanode}^Ta; \quad \sigma^2_{\unode} = \E{\taunode}\E{\taulayer}\sigma_{\betanode}^{2^T}a^2,
\end{split} 
\label{eq:lrpm}
\end{equation}
where $a$ is the input to layer $l$, $\mu_{\betanode}$ and $\sigma^2_{\betanode}$ are the means and variances of the variational posterior over weights incident into node $k$, and $a^2$ denotes a point wise squaring of the input a. Since, the variational posteriors of $\taunode$ and $\taulayer$ are restricted to the log-Normal family, it follows that, $\E{\taunode} = \text{exp}\{\mu_{\taunode} + 0.5 * \sigma^2_{\taunode}\}$, $\E{\taulayer} = \text{exp}\{\mu_{\taulayer} + 0.5 * \sigma^2_{\taulayer}\}$. 

\paragraph{Algorithm} We now have all the tools necessary for optimizing Equation~\ref{eq:elbo}.  By recursively sampling from the variational posterior of Equation~\ref{eq:lrpm} for each layer of the network, we are able to forward propagate information through the network. Owing to the reparameterizations (Equation~\ref{eq:reparam}), we are also able to differentiate through the sampling process and use reverse mode automatic differentiation tools~\cite{DMaclaurin15} to compute the relevant gradients. With the gradients in hand, we optimize $\elbo(\phi)$  with respect to the variational weights $\phi_{\beta}$, per-unit scales $\phi_{\taunode}$, per-layer scales $ \phi_{\taulayer}$, and the variational scale for the output layer weights, $\phi_\kappa$ using Adam~\cite{DKingma2014adam}. Conditioned on these, the optimal variational posteriors of the auxiliary variables $\lambdalayer$, $\lambdanode$, and $\lambdakappa$ follow Inverse Gamma distributions. Fixed point updates that maximize $\elbo(\phi)$ with respect to $\phi_{\lambdalayer}, \phi_{\lambdanode}, \phi_{\lambdakappa}$, holding the other variational parameters fixed are available. The overall algorithm,  involves cycling between gradient and fixed point updates to maximize the ELBO in a coordinate ascent fashion.

\section{Related Work}
Early work on Bayesian neural networks can be traced back
to~\cite{WBuntine91, DMackay92, RNeal93}. These early approaches relied on Laplace approximation or Markov Chain Monte Carlo (MCMC) for inference. They do not scale well to modern architectures or the large datasets required to learn them. Recent advances in stochastic variational methods~\cite{CBlundell15, DRezende14}, black-box variational and alpha-divergence
minimization~\cite{MHLobato16, RRanganath14}, and probabilistic
backpropagation~\cite{MHLobato15} have reinvigorated
interest in BNNs by allowing inference to scale to larger architectures
and larger datasets.

Work on learning structure in BNNs remains relatively nascent. In~\cite{RAdams10} the authors use a cascaded Indian buffet process to learn the structure of sigmoidal belief networks. While interesting, their approach appears susceptible to poor local optima and their proposed Markov Chain Monte Carlo based inference does not scale well. More recently, \cite{CBlundell15} introduce a mixture-of-Gaussians prior on the weights, with one mixture tightly concentrated around zero, thus approximating a spike and slab prior over weights. Their goal of turning off edges is very different than our approach, which performs model selection over the appropriate number of nodes.
Further, our proposed Horseshoe prior can be seen as an extension of their work, where we employ an infinite scale mixture-of-Gaussians. Beyond providing stronger sparsity, this is attractive because it obviates the need to  directly specify the mixture component variances or the mixing proportion as is required by the prior proposed in~\cite{CBlundell15}. Only the prior scales of the variances needs to be specified and in our experiments, we found results to be relatively robust to the values of these scale hyper-parameters. Recent work~\cite{Juho17} indicates that further gains may be possible by a more careful tuning of the scale parameters. Others~\cite{DKingma15, YGal16Dropout} have noticed connections between Dropout~\cite{NSrivastava14} and approximate variational inference. In particular, \cite{DMolchanov17} show that the interpretation of Gaussian dropout as performing variational inference in a network with log uniform priors over weights leads to sparsity in weights. This is an interesting but orthogonal approach, wherein sparsity stems from variational optimization instead of the prior.

There also exists work on learning structure in non-Bayesian neural
networks. Early work~\cite{lecun1990optimal, hassibi1993optimal}
pruned networks by analyzing second-order derivatives of the
objectives. More recently, \cite{wen2016learning} describe
applications of structured sparsity not only for optimizing filters
and layers but also computation time.  Closest to our work in spirit,
\cite{ochiai2016automatic}, \cite{scardapane2017group} and
\cite{murray2015auto} who use group sparsity to prune groups of
weights---e.g. weights incident to a node. However, these approaches
don't model the uncertainty in weights and provide uniform shrinkage
to all parameters. Our horseshoe prior approach similarly
provides group shrinkage while still allowing large
weights for groups that are active.

\section{Experiments}
\label{sec:experiments}
In this section, we present experiments that evaluate various aspects of the proposed Bayesian neural network with horseshoe priors (HS-BNN). We begin with experiments on synthetic data that showcase the model's ability to guard against under fitting and recover the underlying model. We then proceed to benchmark performance on standard regression and classification tasks. For the regression problems we use Gaussian likelihoods with an unknown precision $\gamma$, $p(y_n \mid f(\W, x_n), \gamma) = \normal(y_n \mid f(\W, x_n), \gamma^{-1})$. We place a vague prior on the precision,$\gamma \sim \text{Gamma}(6, 6)$ and approximate the posterior over $\gamma$ using a Gamma distribution. The corresponding variational parameters are learned via a gradient update during learning. We use a Categorical likelihood for the classification problems. In a preliminary study, we found larger mini-batch sizes improved performance, and in all experiments we use a batch size of $512$. The hyper parameters $\bnode$ and $\blayer$ are both set to one. 
\subsection{Experiments on simulated data}
\begin{figure*}[t]
\includegraphics[width=1\textwidth]{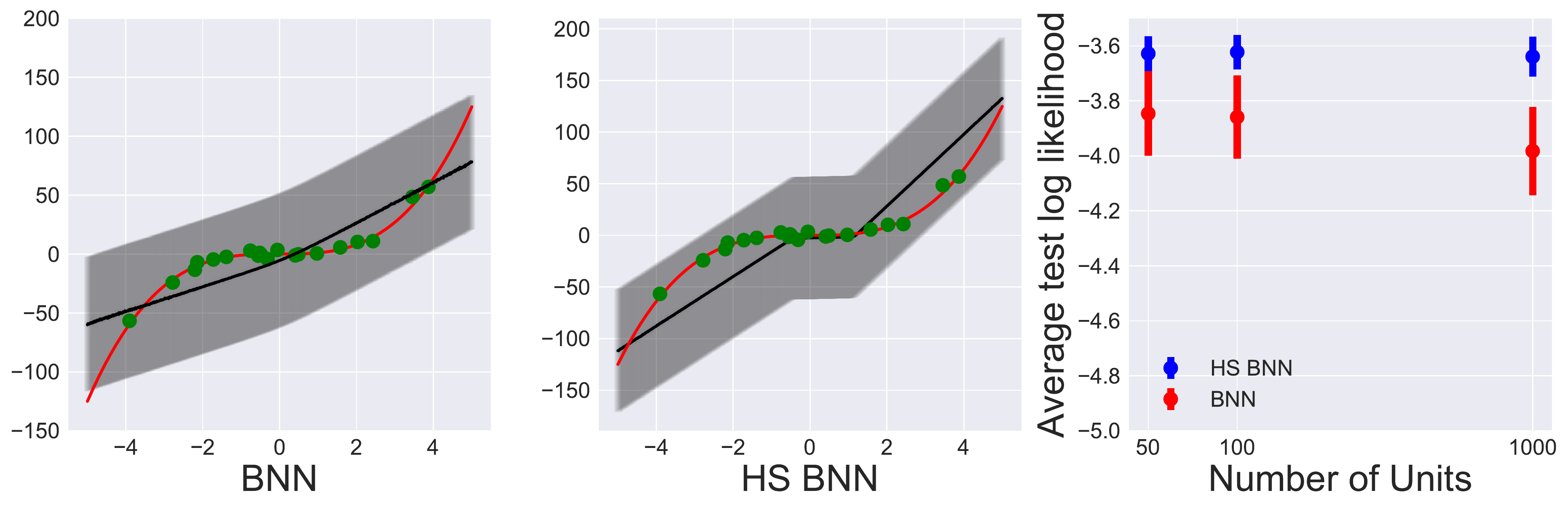}\\
\includegraphics[width=0.25\textwidth]{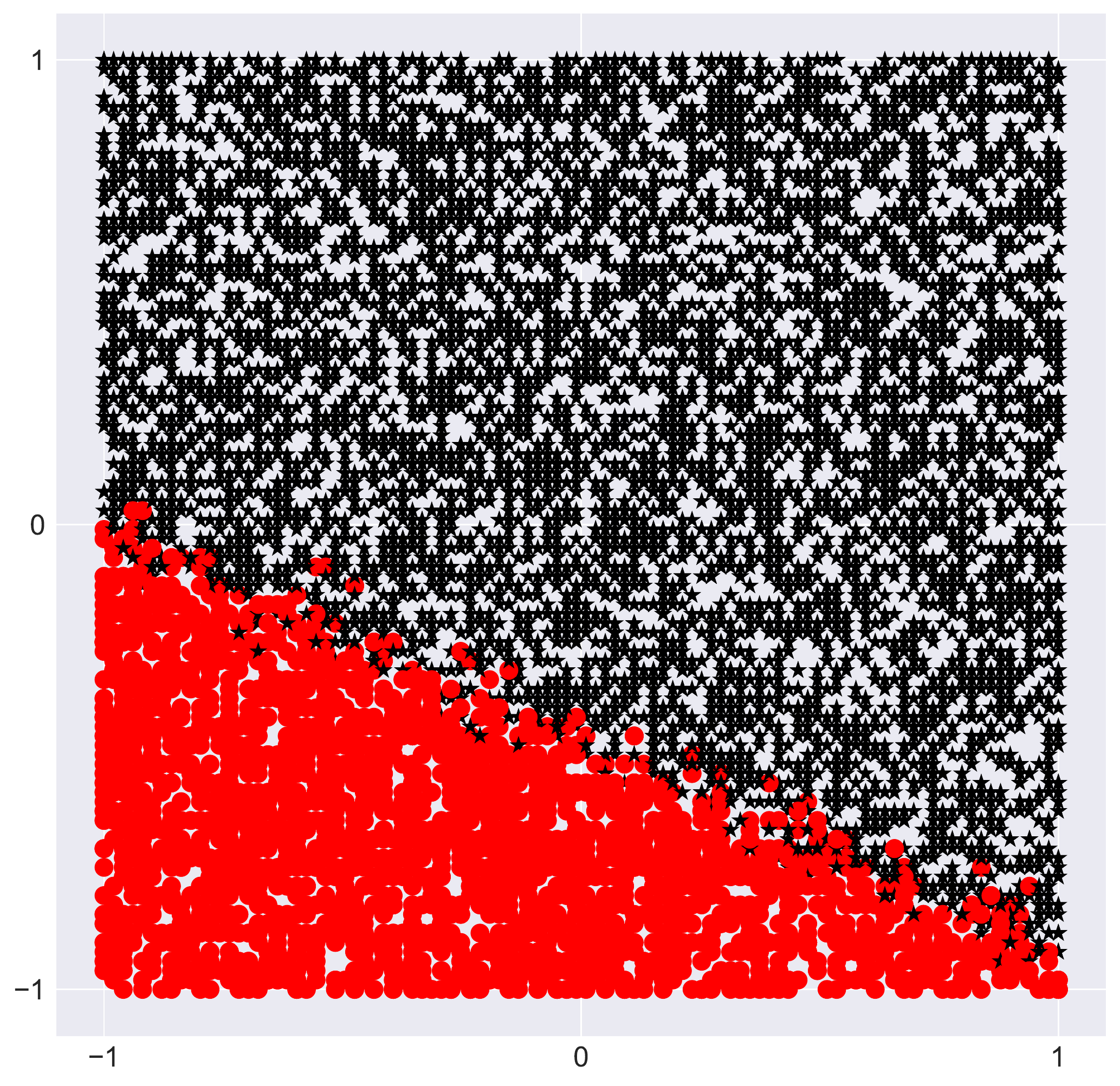}
\includegraphics[width=0.75\textwidth]{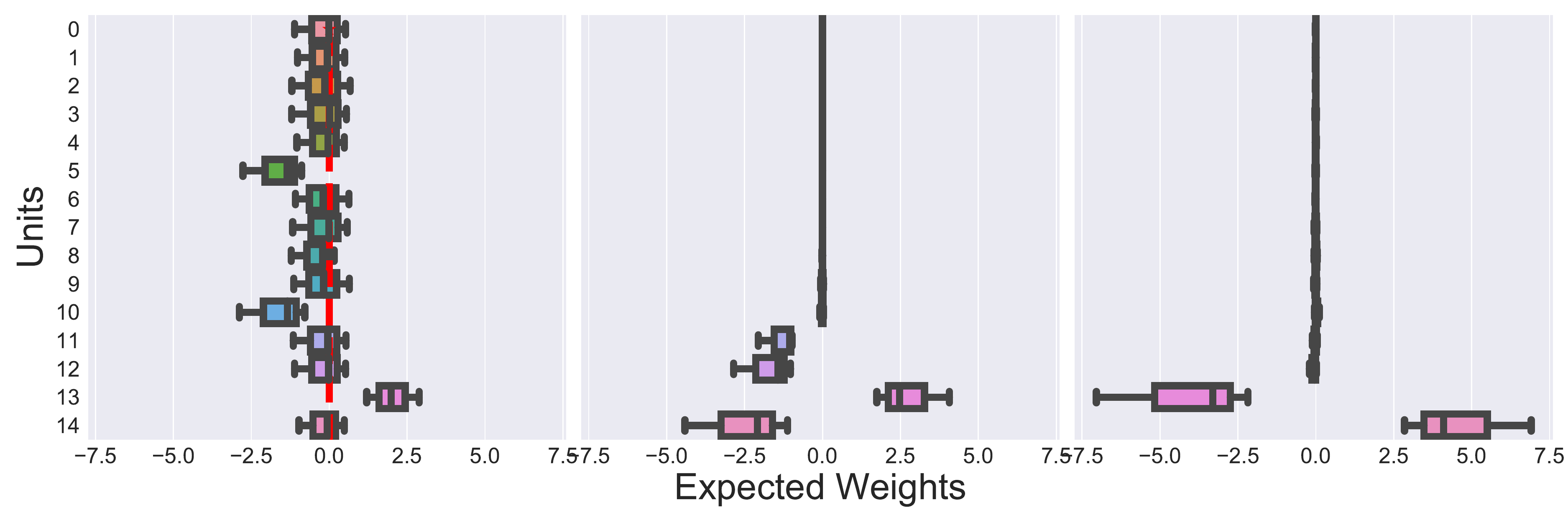}
\caption{\small{TOP: \textbf{Nonlinear one dimensional regression} $y_n = x_n^3 + \epsilon_n$, $\epsilon_n\sim \normal(0, 9)$. The green dots indicate training and test points and the red line visualizes the noise free cubic function and the black lines indicate the predicted means. Left: BNN mean $\pm 3$ standard deviations. Center: HS-BNN mean $\pm 3$ standard deviations. Right: Average predictive log likelihood for single layer networks with 10, 100 and 1000 units. The standard Gaussian prior BNN's predictive performance deteriorates with increasing capacity. HS-BNN is more robust. BOTTOM: \textbf{Non-centered parameterization} is essential for robust inference. From left to right, A synthetic nearly linear classification problem generated from sampling a 2-2-1 network, the two classes are displayed in red and black. Expected weights inferred with a Bayesian neural network with Gaussian priors on weights. Expected weights recovered with a centered horseshoe parameterization. Non-centered horseshoe parameterization. The boxplots display the distribution of expected weights incident onto a hidden unit.}}	
\label{fig:add_cap}
\end{figure*}

\paragraph{Robustness to under-fitting} We begin with a one-dimensional non linear regression problem shown in Figure~\ref{fig:add_cap}. To explore the effect of additional modeling capacity on performance, we sample twenty points uniformly at random in the interval $[-4, +4]$ from the function $y_n = x_n^3 + \epsilon, \epsilon \sim \normal(0,9)$ and train single layer Bayesian neural networks with $50$, $100$ and $1000$ units each. We compare HS-BNN against a BNN with Gaussian priors on weights, $w_{ij,l} \sim \normal(0, \kappa), \kappa\sim \halfcauchy(0, 5)$, training both for a $1000$ iterations. The performance of the BNN with Gaussian priors quickly deteriorates with increasing capacity as a result of under fitting the limited amount of training data. In contrast, HS-BNN by pruning away additional capacity is more robust to model misspecification showing only a marginal drop in predictive performance with increasing number of units.
\paragraph{Non-centered parameterization}
Next, we explore the benefits of the non-centered parameterization. We consider a simple two dimensional classification problem generated by sampling data uniformly at random from $[-1, +1] \times [-1, +1]$ and using a 2-2-1 network, whose parameters are known \emph{a-priori} to generate the class labels. We train three Bayesian neural networks with a $15$ unit layer on this data, with Gaussian priors, with horseshoe priors but employing a centered parameterization, and with the non-centered horseshoe prior. Each model is trained till convergence. We find that all three models are able to easily fit the data and provide high predictive accuracy. However, the structure learned by the three models are very different. In Figure~\ref{fig:add_cap}  we visualize the distribution of weights incident onto a unit. Unsurprisingly, the BNN with Gaussian priors does not exhibit sparsity. In contrast, models employing the horseshoe prior are able to prune units away by setting all incident weights to tiny values. It is interesting to note that even for this highly stylized example the centered parameterization struggles to recover the true structure of the underlying network. The non-centered parameterization however does significantly better and prunes away all but two units. Further experiments provided in the supplement demonstrate the same effect for wider 100 unit networks. The non-centered parameterized model is again able to recover the two active units.
\subsection{Classification and Regression experiments}
\begin{figure*}[t]
\begin{center}
\includegraphics[width=0.99\textwidth]{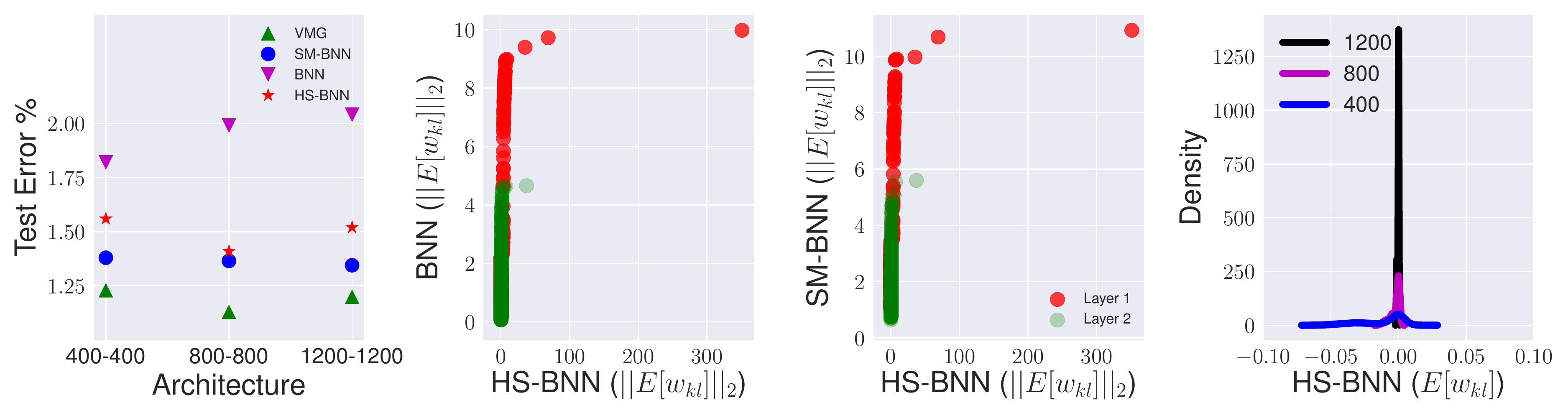}
\includegraphics[width=0.485\textwidth]{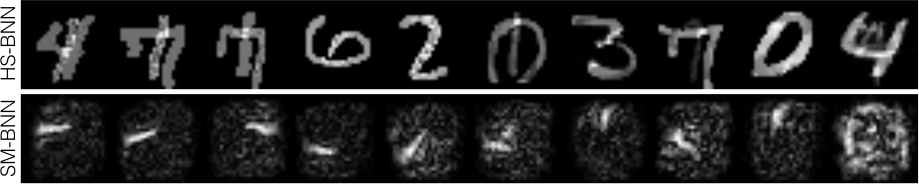}
\includegraphics[width=0.485\textwidth]{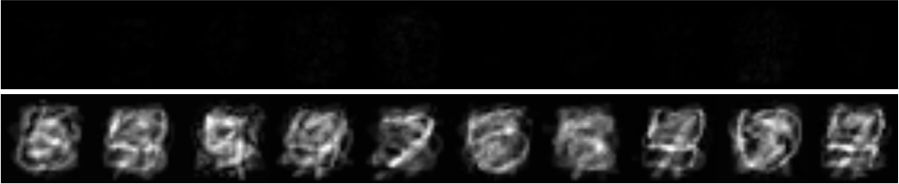}
\end{center}
\caption{\small{MNIST experiments. TOP: \emph{From left to right}, Test error rates for different architectures and methods. The right two plots compare the sparsity of the solutions found by HS-BNN, SM-BNN and BNN. For the 1200-1200 network, we compare the expected node weight vectors inferred under the different models. We sort the recovered weight vectors $\E{\wnode}$ based on their 2-norm and compare them via scatter plots. Each circle corresponds to one of the $1200$ weight node vectors. Compared to competing methods a large number of weight node vectors are zeroed out, while a small number escapes un-shrunk for HS-BNN. The rightmost plot shows the density of the unit with the lowest norm from the the three architectures. BOTTOM: $\E{\wnode}$ for the first layer. The left and right columns visualize the ten units with the largest and the smallest norms.}}
\label{fig:mnist_results}
\end{figure*}
We benchmark classification performance on the MNIST dataset. Additional experiments on a gesture recognition task are available in the supplement. We compare HS-BNN against the variational matrix Gaussian (VMG)~\cite{CLouizos16}, a BNN with a two-component scale mixture (SM-BNN) prior on weights proposed in~\cite{CBlundell15} and a BNN with Gaussian prior (BNN) on weights. VMG uses a structured variational approximation, while the other approaches all use fully factorized approximations and differ only in the type of prior used. These approaches constitute the \emph{state-of-the-art} in variational learning for Bayesian neural networks.   
\paragraph{MNIST} We preprocessed the images in the MNIST digits dataset by dividing the pixel values by 126. We explored networks with varying widths and depths all employing rectified linear units. For HS-BNN we used Adam with a learning rate of $0.005$ and $500$ epochs. We did not use a validation set to monitor validation performance or tune hyper-parameters. We used the parameter settings recommended in the original papers for the competing methods.  Figure~\ref{fig:mnist_results} summarizes our findings. We showcase results for three architectures with two hidden layers each containing $400$, $800$ and $1200$ rectified linear hidden units. Across architectures, we find our performance to be significantly better than BNN, comparable to SM-BNN, and worse than VMG. The poor performance with respect to VMG likely stems from the structured matrix variate variational approximation employed by VMG. 

More interestingly, we clearly see the sparsity inducting effects of the horseshoe prior. Recall that under the horseshoe prior, $\wnode \sim \normal(0, \taunode^2\taulayer^2\eye)$. As the scales $\taunode\taulayer$ tend to zero the corresponding units (and all incident weights) are pruned away. SM-BNN also encourages sparsity, but on weights not nodes. Further, the horseshoe prior with its thicker tails and taller spike at origin encourages stronger sparsity. To see this we compared the 2-norms of the inferred expected weight node vectors $\E{\wnode}$ found by SM-BNN and HS-BNN (Figure~\ref{fig:mnist_results}). For HS-BNN the inferred scales are tiny for most units, with a few notable outliers that escape un-shrunk. This causes the corresponding weight vectors to be zero for the majority of units, suggesting that the model is able to effectively ``turn off'' extra capacity. In contrast, the weight node vectors recovered by SM-BNN (and BNN) are less tightly concentrated at zero. We also plot the density of $\E{\wnode}$ with the smallest norm in each of the three architectures. Note that with increasing architecture size (modeling capacity) the density peaks more strongly at zero, suggesting that the model is more confident in turning off the unit and not use the extra modeling capacity.
To further explore the implications of node versus weight sparsity, we visualize $\E{\wnode}$ learned by SM-BNN and HS-BNN in Figure~\ref{fig:mnist_results}. Weight sparsity in SM-BNN encourages fundamentally different filters that pick up edges at different orientations. In contrast, HS-BNN's node sparsity encourages filters that correspond to digits or superpositions of digits and may lead to more interpretable networks. Stronger sparsity afforded by the horseshoe is again evident when visualizing filters with the lowest norms.  HS-BNN filters are nearly all black when scaled with respect to the SM-BNN filters. 
\\\\
\textbf{Regression} We also compare the performance of our model on regression datasets from the UCI repository. We follow the experimental protocol proposed in~\cite{MHLobato15, CLouizos16} and train a single hidden layer network with $50$ rectified linear units for all but the larger ``Protein'' and ``Year'' datasets for which we train a $100$ unit network. For the smaller datasets we train on a randomly subsampled $90\%$ subset and evaluate on the remainder and repeat this process $20$ times. For ``Protein'' we perform 5 replications and for ``Year'' we evaluate on a single split. Here, we only benchmark against VMG, which has previously been shown to outperform alternatives~\cite{CLouizos16}. Table~\ref{tab:regression} summarizes our results. Despite our fully factorized variational approximation we remain competitive with VMG in terms of both root mean squared error (RMSE) and predictive log likelihoods and even outperform it on some datasets. A more careful selection of the scale hyper-parameters~\cite{Juho17}, and the use of structured variational approximations similar to VMG will likely help improve results further and constitute interesting directions of future work. 
\begin{table*}
\begin{center}
\resizebox{0.95\textwidth}{!}{
\begin{tabular}{ l  l  l  l l l}
Dataset & N(d) & VMG(RMSE) & HS-BNN (RMSE) & VMG(Test ll) & HS-BNN(Test ll)\\
\hline
 Boston & 506 (13) & $\bf{2.72\pm0.13}$ & $3.32\pm 0.66$ & $\bf{-2.48\pm 0.66}$ & $-2.54\pm 0.15$ \\
 Concrete & 1030 (8) & $\bf{4.88\pm0.12}$ & $5.66 \pm 0.41$ & $\bf{-3.01\pm 0.66}$ & $-3.09\pm0.06$ \\
 Energy & 768 (8) & $\bf{0.54 \pm 0.02}$ & $1.99\pm0.34$ & $\bf{-1.06\pm0.03}$ & $-2.66 \pm 0.13$\\
 Kin8nm & 8192 (8) & $\bf{0.08 \pm 0.00}$ & $\bf{0.08 \pm 0.00}$ & $+1.10\pm 0.01$ & $\bf{+1.12\pm 0.03}$ \\
 Naval & 11,934 (16) & $\bf{0.00 \pm 0.00}$ & $\bf{0.00 \pm 0.00}$ & $+2.46\pm 0.00$ & $\bf{+5.52\pm 0.10}$\\
 Power Plant & 9568 (4) & $4.04 \pm 0.04$ & $\bf{4.03 \pm 0.15}$ & $-2.82\pm 0.01$ & $\bf{-2.81 \pm 0.03}$ \\
 Protein & 45.730 (9) & $\bf{4.18 \pm 0.02}$ & $4.39\pm 0.04$& $\bf{-2.88\pm 0.00}$ & $-2.89\pm 0.01$ \\
 Wine & 1599 (11) & $\bf{0.63 \pm 0.01}$ & $\bf{0.63\pm 0.04}$ &$\bf{-0.95\pm 0.01}$ & $\bf{-0.95\pm 0.05}$ \\
 Yacht & 308 (6) & $\bf{0.70 \pm 0.05}$ & $1.58\pm 0.23$ & $\bf{-1.30 \pm 0.02}$ & $-2.33 \pm 0.01$ \\
 Year & 515,345 (90) & $\bf{8.82\pm\text{NA}}$ & $9.26\pm \text{NA}$  &$\bf{-3.60\pm\text{NA}}$& $-3.63\pm\text{NA}$\\
 \hline
 \end{tabular}}
  \end{center}
\caption{\small{UCI Regression results. HS-BNN and VMG are compared.}}
\label{tab:regression}
\end{table*}

\section{Discussion and Conclusion}

In Section~\ref{sec:experiments}, we demonstrated that a properly
parameterized horseshoe prior on the scales of the weights incident to
each node is a computationally efficient tool for model selection in
Bayesian neural networks.  Decomposing the horseshoe prior into
inverse gamma distributions and using a non-centered representation ensured a degree of robustness to poor local optima.
While we have seen that the horseshoe prior is an effective tool for
model selection, one might wonder about more common alternatives. We lay out a few obvious choices and contrast
their deficiencies. One starting point is to observe that a node can
be pruned if all its incident weights are zero (in this case, it can
only pass on the same bias term to the rest of the network).  Such
sparsity can be encouraged by a simple exponential prior on the weight
scale, but without heavy tails all scales are forced artificially low
and prediction suffers and has been noted in the context of learning sparse neural networks~\cite{DMolchanov17, wen2016learning}. In contrast, simply using a heavy-tail prior
on the scale parameter, such as a half-Cauchy, will not apply any
pressure to set small scales to zero, and we will not have sparsity.
Both the shrinkage to zero and the heavy tails of the horseshoe prior
are necessary to get the model selection that we require.  And
importantly, using a continuous prior with the appropriate statistical
properties is simple to incorporate with existing inference, unlike an
explicit spike and slab model.  Another alternative is to
observe that a node can be pruned if the product $z \cdot w$ is nearly
constant for all inputs $z$---having small weights is sufficient to
achieve this property; weights $w$ that are orthogonal to the
variation in $z$ is another.  Thus, instead of putting a prior over
the scale of $w$, one could put a prior over the scale of the
variation in $z \cdot w$.  While we believe this is more general, we
found that such a formulation has many more local optima and thus
harder to optimize.  



{\small
\bibliographystyle{ieee}
\bibliography{egbib}

\begin{thebibliography}{10}\itemsep=-1pt

\bibitem{RAdams10}
R.~P. Adams, H.~M. Wallach, and Z.~Ghahramani.
\newblock Learning the structure of deep sparse graphical models.
\newblock In {\em AISTATS}, 2010.

\bibitem{MBetancourt15}
M.~Betancourt and M.~Girolami.
\newblock Hamiltonian monte carlo for hierarchical models.
\newblock {\em Current trends in Bayesian methodology with applications},
  79:30, 2015.

\bibitem{CBlundell15}
C.~Blundell, J.~Cornebise, K.~Kavukcuoglu, and D.~Wierstra.
\newblock Weight uncertainty in neural networks.
\newblock In {\em ICML}, pages 1613--1622, 2015.

\bibitem{WBuntine91}
W.~L. Buntine and A.~S. Weigend.
\newblock Bayesian back-propagation.
\newblock {\em Complex systems}, 5(6):603--643, 1991.

\bibitem{CCarvalho09}
C.~M. Carvalho, N.~G. Polson, and J.~G. Scott.
\newblock Handling sparsity via the horseshoe.
\newblock In {\em AISTATS}, 2009.

\bibitem{YGal16Dropout}
Y.~Gal and Z.~Ghahramani.
\newblock Dropout as a {B}ayesian approximation: Representing model uncertainty
  in deep learning.
\newblock In {\em ICML}, 2016.

\bibitem{YGal16}
Y.~Gal and Z.~Ghahramani.
\newblock A theoretically grounded application of dropout in recurrent neural
  networks.
\newblock In {\em NIPS}, 2016.

\bibitem{YGal16Active}
Y.~Gal, R.~Islam, and Z.~Ghahramani.
\newblock Deep {B}ayesian active learning with image data.
\newblock In {\em Bayesian Deep Learning workshop, NIPS}, 2016.

\bibitem{hassibi1993optimal}
B.~Hassibi, D.~G. Stork, and G.~J. Wolff.
\newblock Optimal brain surgeon and general network pruning.
\newblock In {\em Neural Networks, 1993., IEEE Intl. Conf. on}, pages 293--299.
  IEEE, 1993.

\bibitem{MHLobato16}
J.~Hernandez-Lobato, Y.~Li, M.~Rowland, T.~Bui, D.~Hern{\'a}ndez-Lobato, and
  R.~Turner.
\newblock Black-box alpha divergence minimization.
\newblock In {\em ICML}, pages 1511--1520, 2016.

\bibitem{MHLobato15}
J.~M. Hern\'{a}ndez-Lobato and R.~P. Adams.
\newblock Probabilistic backpropagation for scalable learning of bayesian
  neural networks.
\newblock In {\em ICML}, 2015.

\bibitem{JIngraham16}
J.~B. Ingraham and D.~S. Marks.
\newblock Bayesian sparsity for intractable distributions.
\newblock {\em arXiv:1602.03807}, 2016.

\bibitem{AJoshi17}
A.~Joshi, S.~Ghosh, M.~Betke, S.~Sclaroff, and H.~Pfister.
\newblock Personalizing gesture recognition using hierarchical bayesian neural
  networks.
\newblock In {\em CVPR}, 2017.

\bibitem{DKingma2014adam}
D.~Kingma and J.~Ba.
\newblock Adam: A method for stochastic optimization.
\newblock {\em arXiv:1412.6980}, 2014.

\bibitem{DKingma15}
D.~P. Kingma, T.~Salimans, and M.~Welling.
\newblock Variational dropout and the local reparameterization trick.
\newblock In {\em NIPS}, 2015.

\bibitem{DKingma14}
D.~P. Kingma and M.~Welling.
\newblock Stochastic gradient {VB} and the variational auto-encoder.
\newblock In {\em ICLR}, 2014.

\bibitem{lecun1990optimal}
Y.~LeCun, J.~S. Denker, and S.~A. Solla.
\newblock Optimal brain damage.
\newblock In {\em NIPS}, pages 598--605, 1990.

\bibitem{CLouizos16}
C.~Louizos and M.~Welling.
\newblock Structured and efficient variational deep learning with matrix
  {G}aussian posteriors.
\newblock In {\em ICML}, pages 1708--1716, 2016.

\bibitem{DMackay92}
D.~J. MacKay.
\newblock A practical {Bayesian} framework for backpropagation networks.
\newblock {\em Neural computation}, 4(3):448--472, 1992.

\bibitem{DMaclaurin15}
D.~Maclaurin, D.~Duvenaud, and R.~P. Adams.
\newblock Autograd: Effortless gradients in numpy.
\newblock In {\em ICML AutoML Workshop}, 2015.

\bibitem{DMolchanov17}
D.~Molchanov, A.~Ashukha, and D.~Vetrov.
\newblock Variational dropout sparsifies deep neural networks.
\newblock {\em arXiv:1701.05369}, 2017.

\bibitem{murray2015auto}
K.~Murray and D.~Chiang.
\newblock Auto-sizing neural networks: With applications to n-gram language
  models.
\newblock {\em arXiv:1508.05051}, 2015.

\bibitem{RNeal93}
R.~M. Neal.
\newblock Bayesian learning via stochastic dynamics.
\newblock In {\em NIPS}, 1993.

\bibitem{ochiai2016automatic}
T.~Ochiai, S.~Matsuda, H.~Watanabe, and S.~Katagiri.
\newblock Automatic node selection for deep neural networks using group lasso
  regularization.
\newblock {\em arXiv:1611.05527}, 2016.

\bibitem{Juho17}
J.~Piironen and A.~Vehtari.
\newblock On the hyperprior choice for the global shrinkage parameter in the
  horseshoe prior.
\newblock {\em AISTATS}, 2017.

\bibitem{RRanganath14}
R.~Ranganath, S.~Gerrish, and D.~M. Blei.
\newblock Black box variational inference.
\newblock In {\em AISTATS}, pages 814--822, 2014.

\bibitem{DRezende14}
D.~J. Rezende, S.~Mohamed, and D.~Wierstra.
\newblock Stochastic backpropagation and approximate inference in deep
  generative models.
\newblock In {\em ICML}, pages 1278--1286, 2014.

\bibitem{scardapane2017group}
S.~Scardapane, D.~Comminiello, A.~Hussain, and A.~Uncini.
\newblock Group sparse regularization for deep neural networks.
\newblock {\em Neurocomputing}, 241:81--89, 2017.

\bibitem{song2011tracking}
Y.~Song, D.~Demirdjian, and R.~Davis.
\newblock Tracking body and hands for gesture recognition: Natops aircraft
  handling signals database.
\newblock In {\em Automatic Face \& Gesture Recognition and Workshops (FG
  2011), 2011 IEEE International Conference on}, pages 500--506. IEEE, 2011.

\bibitem{NSrivastava14}
N.~Srivastava, G.~E. Hinton, A.~Krizhevsky, I.~Sutskever, and R.~Salakhutdinov.
\newblock Dropout: a simple way to prevent neural networks from overfitting.
\newblock {\em JMLR}, 15(1):1929--1958, 2014.

\bibitem{MTitsias14}
M.~Titsias and M.~L{\'a}zaro-gredilla.
\newblock Doubly stochastic variational {Bayes} for non-conjugate inference.
\newblock In {\em ICML}, pages 1971--1979, 2014.

\bibitem{MWand11}
M.~P. Wand, J.~T. Ormerod, S.~A. Padoan, R.~Fuhrwirth, et~al.
\newblock Mean field variational {B}ayes for elaborate distributions.
\newblock {\em Bayesian Analysis}, 6(4):847--900, 2011.

\bibitem{wen2016learning}
W.~Wen, C.~Wu, Y.~Wang, Y.~Chen, and H.~Li.
\newblock Learning structured sparsity in deep neural networks.
\newblock In {\em NIPS}, pages 2074--2082, 2016.

\bibitem{RWilliams92}
R.~J. Williams.
\newblock Simple statistical gradient-following algorithms for connectionist
  reinforcement learning.
\newblock {\em Machine learning}, 8(3-4):229--256, 1992.

\end{thebibliography}
}
\newpage
\appendix
\section {Fixed point updates}
The ELBO corresponding to the non-centered HS model is,
\begin{equation}
\begin{split}
	\elbo(\phi) &= \E{\text{ln }\invgamma(\kappa \mid 1/2, 1/\lambdakappa)} + \E{\text{ln }\invgamma(\lambdakappa \mid 1/2, 1/\bkappa^2)}  \\
	&+\sum_n \E{\text{ln } p(y_n \mid \beta, \mathcal{T}, \kappa, x_n)} \\
	&+\sum_{l=1}^{L-1}\sum_{k=1}^{K_L} \E{\text{ln }\invgamma(\taunode \mid 1/2, 1/\lambdanode)}
	 \E{\text{ln }\invgamma(\lambdanode \mid 1/2, 1/\bnode^2)}\\
	  &+\sum_{l=1}^{L-1} \E{\text{ln }\invgamma(\taulayer\mid 1/2, 1/\lambdalayer)} + \E{\text{ln }\invgamma(\lambdalayer\mid 1/2, 1/\blayer^2)}\\
	   &+\sum_{l=1}^{L-1}\sum_{k=1}^{K_l}\E{\text{ln }\normal(\beta_{kl}\mid 0, \eye) } + \sum_{k=1}^{K_L} \E{\text{ln }\normal(\beta_{kL}\mid 0, \eye )} + \ent{q(\theta\mid \phi)}.
\end{split}
\end{equation}
With our choices of the variational approximating families, all the entropies are available in closed form. We rely on a Monte-Carlo estimates to evaluate the expectation involving the likelihood $\E{\text{ln } p(y_n \mid \beta, \mathcal{T}, \kappa, x_n)}$. 

The auxiliary variables $\lambdakappa$, $\lambdalayer$ and $\lambdalayer$ all follow inverse Gamma distributions. Here we derive for $\lambdanode$, the others follow analogously.
Consider,
\begin{equation}
\begin{split}
	\text{ln } q(\lambdanode) &\propto \Ewrt{-q_{\lambdanode}}{\text{ln }\invgamma(\taunode \mid 1/2, 1/\lambdanode)} + \Ewrt{-q_{\lambdanode}}{\text{ln }\invgamma(\lambdanode \mid 1/2, 1/\bnode^2)}, \\
	&\propto (-1/2 -1/2 -1)\text{ln } \lambdanode - (\E{1/\taunode} + 1/\bnode^2)(1/\lambdanode),
	\end{split}
\end{equation}
from which we see that,
\begin{equation}
\begin{split}
q(\lambdanode) = \invgamma(\lambdanode \mid c, d), \\
c = 1, d = \E{\frac{1}{\taunode}} + \frac{1}{\bnode^2}.	
\end{split}
\end{equation}
Since, $q(\taunode) = \text{ln }\normal(\mu_{\taunode}, \sigma^2_{\taunode})$, it follows that $\E{\frac{1}{\taunode}} = \text{exp}\{-\mu_{\taunode} + 0.5*\sigma^2_{\taunode}\}$. We can thus calculate the necessary fixed point updates for $\lambdanode$ conditioned on $\mu_{\taunode}$ and $\sigma^2_{\taunode}$. Our algorithm uses these fixed point updates given estimates of $\mu_{\taunode}$ and $\sigma^2_{\taunode}$ after each Adam step.

\section{Additional Experiments}
\subsection{Simulated Data}
\begin{figure*}
\includegraphics[width=1\textwidth]{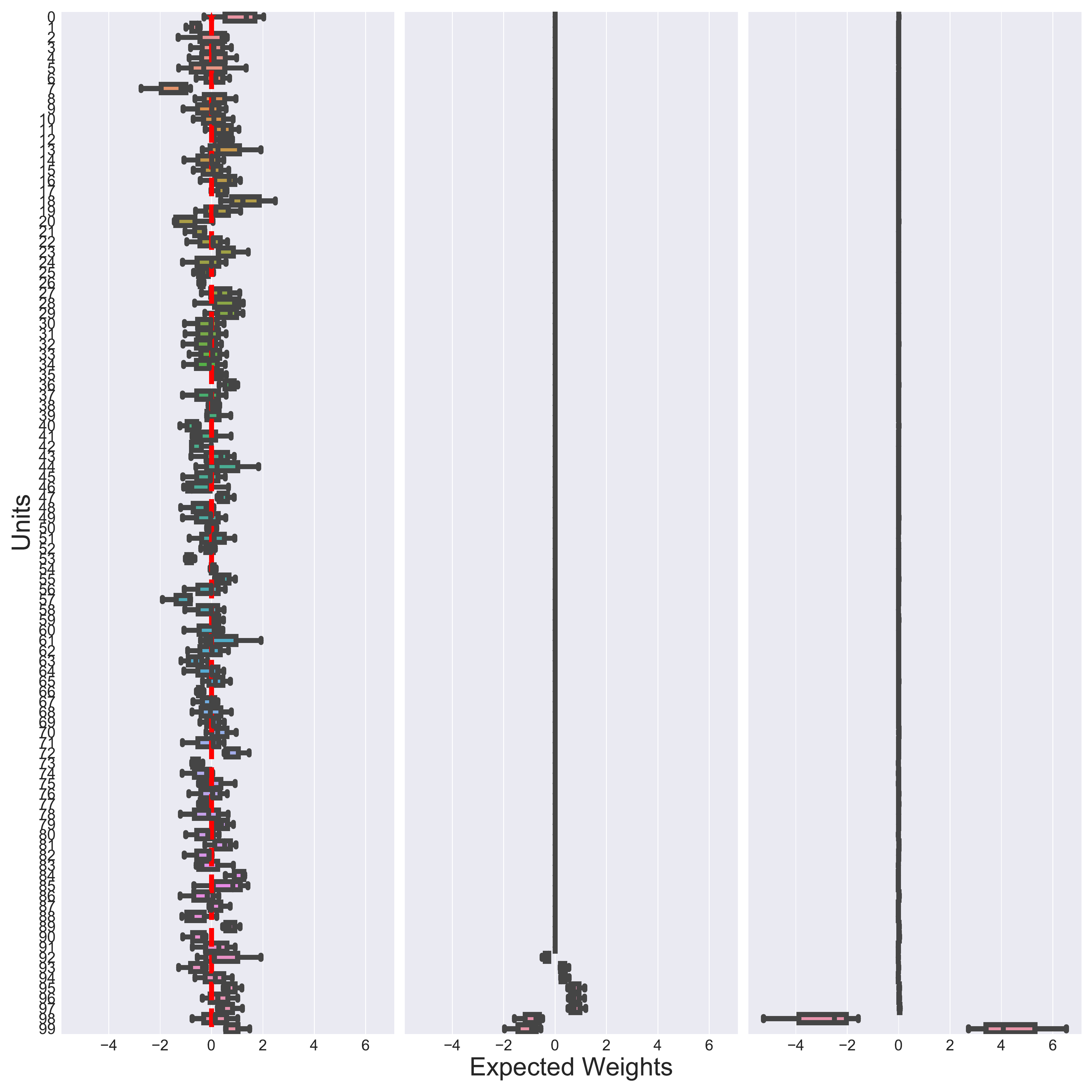}
\caption{Learned sparsity on a synthetic classification problem. We use a single hidden layer network with 100 units. Left: Bayesian neural network with Gaussian. Center: Horseshoe, centered parameterization. Right: Horseshoe, non-centered parameterization.} 
\label{fig:sup1}
\end{figure*}

Here we provide an additional experiment with the data setup in Section 6.2. We use the same linearly separable data, but train larger networks with 100 units each. Figure~\ref{fig:sup1} shows the inferred weights under the different models. Observe that the non-centered HS-BNN is again able to prune away extra capacity and recover two active nodes.

\subsection{Further Exploration of Model Selection Properties}
Here we provide additional results that illustrate the model selection abilities of HS-BNN. First we visualize the norms of inferred node weight vectors  $\E{\wnode}$ found by BNN, SM-BNN and HS-BNN for $400-400$, $800-800$ and $1200-1200$ networks. Note that as we increase capacity the model selection abilities of HS-BNN becomes more obvious and as opposed to the other approaches illustrate clear inflection points and it is evident that the model is using only a fraction of its available capacity.
\begin{figure}
\centering
\includegraphics[width=1\textwidth]{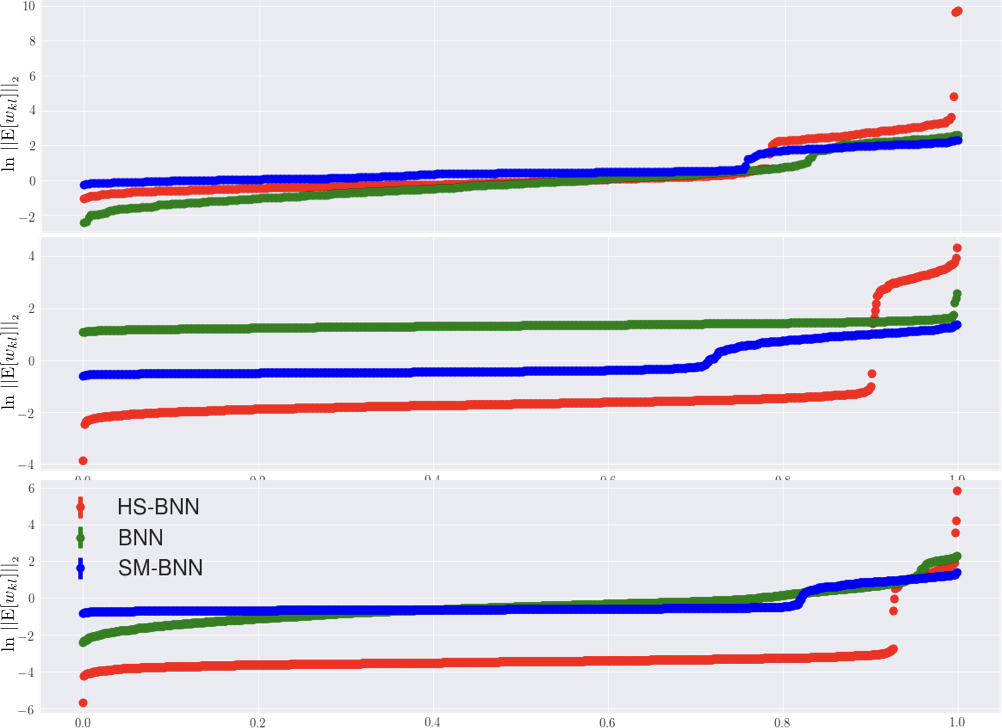}
\caption{Model selection in HS-BNN. From top to bottom, we plot $\text{log }||\E{\wnode}||_2$ for the first layer of a network with 400, 800 and 1200 units. The x-axis has been normalized by the number of units. }	
\end{figure}

As a reference we compare against SM-BNN. We visualize the density of the inferred node weight vectors $\E{\wnode}$ under the two models for networks $400-400$, $800-800$ and $1200-1200$. For each network we show the density of the $5$ units with the smallest norms from either layer. Note that in all three cases HS-BNN produces weights that are more tightly concentrated around zero. Moreover for HS-BNN the concentration around zero becomes sharper with increasing modeling capacity (larger architectures), again indicating that we are pruning away additional capacity.
\begin{figure}
\includegraphics[width=1\textwidth]{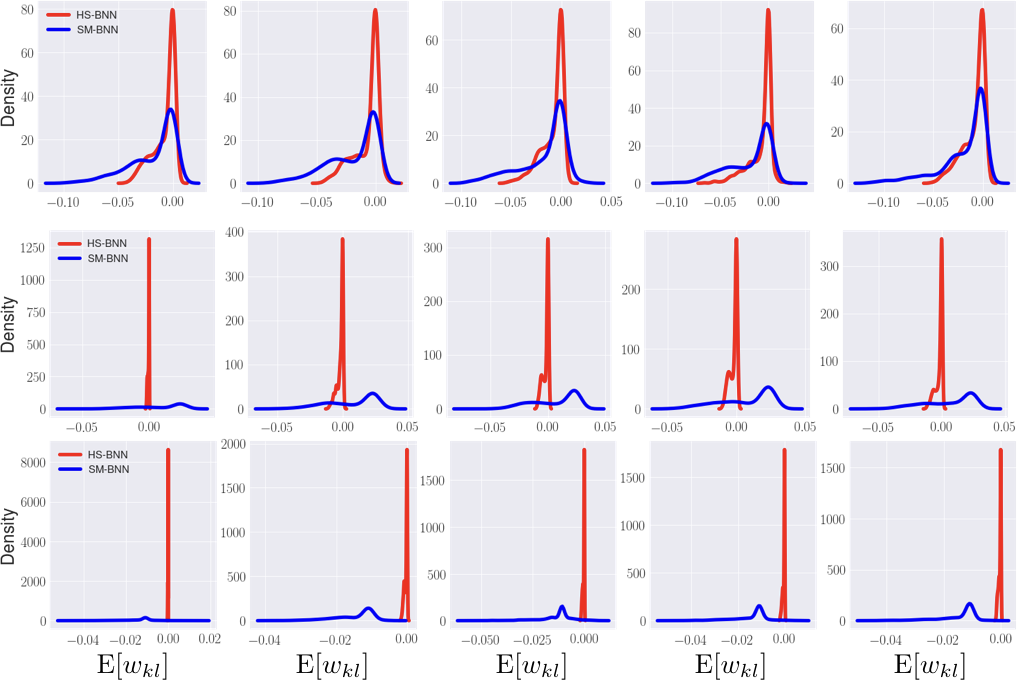}
\caption{Further exploration of sparse solutions found by HS-BNN. Here we provide density plots for the smallest node weight vectors $\wnode$ found by HS-BNN and SM-BNN for the 400-400 (top row), 800-800 (middle row), 1200-1200(bottom row) network. The plots are sorted by 2-norm of $\wnode$, from left to right. }	
\end{figure}

%

\subsection{Gesture Recognition}
\textbf{Gesture recognition}
\begin{figure*}[t]
\begin{center}
\includegraphics[width=1\textwidth]{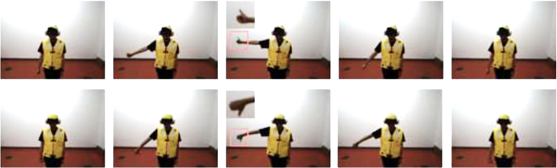}\\
\includegraphics[width=1\textwidth]{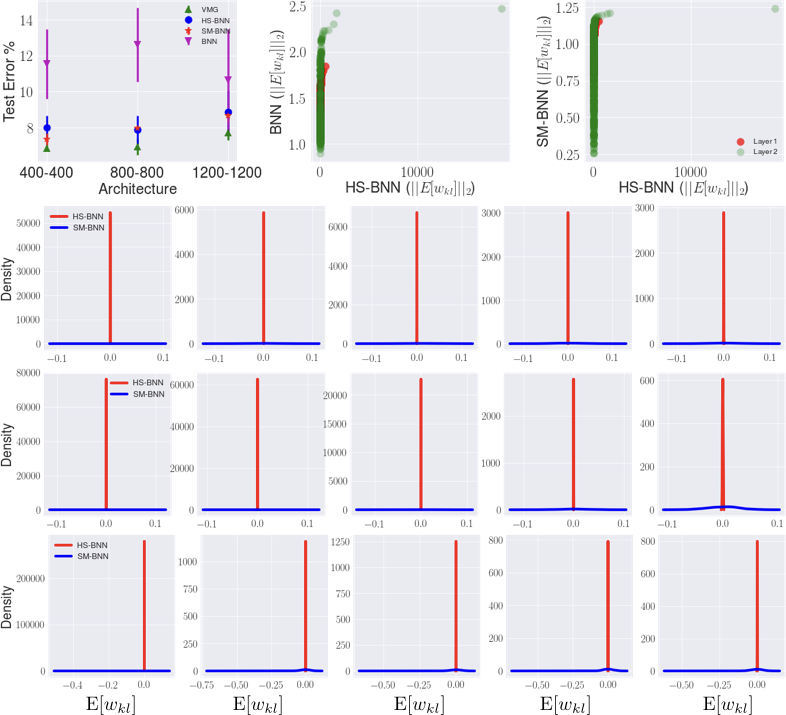}
\end{center}
\caption{Gesture recognition. Top: Example gestures from the NATOPS dataset. Second Row: \emph{Left}: Test error rates averaged over 5 runs achieved by  competing methods. \emph{Right}: Scatter plots of solutions recovered by HS-BNN and competing methods for the 800-800 architecture on one of the five splits. The bottom three rows provide density plots for the smallest node weight vectors $\wnode$ found by HS-BNN and SM-BNN for the 400-400 (top row), 800-800 (middle row), 1200-1200(bottom row) network. The plots are sorted by 2-norm of $\wnode$, from left to right. }	
\label{fig:gesture}
\end{figure*}
We also experimented with a gesture recognition dataset~\cite{song2011tracking} that consists of 24 unique aircraft handling signals performed by 20 different subjects, each for 20 repetitions. The task consists of recognizing these gestures from kinematic, tracking and video data. However, we only use kinematic and tracking data. A couple of example gestures are visualized in Figure~\ref{fig:gesture}. The dataset contains $9600$ gesture examples.

A 12-dimensional vector of body features (angular joint velocities for the right and left elbows and wrists), as well as an 8 dimensional vector of hand features (probability values for hand shapes for the left and right hands) collected by Song et al. ~\cite{song2011tracking} are provided as features for all frames of all videos in the dataset. We additionally used the 20 dimensional per-frame tracking features made available in~\cite{song2011tracking}. We constructed features to represent each gesture by first extracting frames by sampling  uniformly in time and then concatenating the per-frame features of the selected frames to produce 600-dimensional feature vectors. 

This is a much smaller dataset than MNIST and recent work~\cite{AJoshi17} has demonstrated that a BNN with Gaussian priors performs well on this task. Figure~\ref{fig:gesture} compares the performance of HS-BNN with competing methods. We train a two layer HS-BNN with each layer containing 400 units. The error rates reported are a result of averaging over 5 random 75/25 splits of the dataset. Similar to MNIST, HS-BNN significantly outperforms BNN and is competitive with VMG and SM-BNN. We also see strong sparsity, just as in MNIST.

\end{document}